\documentclass[letterpaper, 10 pt, conference]{ieeeconf}

\usepackage{amsmath,amsfonts,bm}

\def\eqref#1{equation~\ref{#1}}
\def\1{\bm{1}}

\DeclareMathAlphabet{\mathsfit}{\encodingdefault}{\sfdefault}{m}{sl}
\SetMathAlphabet{\mathsfit}{bold}{\encodingdefault}{\sfdefault}{bx}{n}

\usepackage{subfigure}
\usepackage{amsmath,amssymb,amsfonts}
\usepackage{graphicx}
\usepackage{textcomp}
\usepackage{xcolor}
\usepackage{hyperref}
\usepackage{url}
\usepackage{adjustbox}
\usepackage{tabu}
\usepackage{booktabs}       % professional-quality tables
\usepackage{wrapfig}
\usepackage{multirow}
\usepackage{subfigure}
\usepackage{rotating} 
\usepackage{adjustbox}
\usepackage{algorithm}
\usepackage{algpseudocode}
\usepackage{caption}
\usepackage{cleveref}
\IEEEoverridecommandlockouts
\title{\LARGE \bf{SplaTraj:} Camera Trajectory Generation \\ with Semantic Gaussian Splatting}

\author{
 Xinyi Liu$^{1,2}$, Tianyi Zhang$^{1}$, Matthew Johnson-Roberson$^{1}$ and Weiming Zhi$^{1}$
\thanks{$^{1}$ Robotics Institute,
        Carnegie Mellon University,
        Pittsburgh, PA, USA
        {\tt\small Email: wzhi@andrew.cmu.edu}}
\thanks{$^{2}$ Joint Department of Biomedical Engineering, University of North Carolina-Chapel Hill, Raleigh, NC,  USA
{\tt\small Email: xinyili@ad.unc.edu}}
}

\begin{document}
\maketitle
\thispagestyle{empty}
\pagestyle{empty}

\begin{abstract}
Many recent developments for robots to represent environments have focused on photorealistic reconstructions. This paper particularly focuses on generating sequences of images from the photorealistic Gaussian Splatting models, that match instructions that are given by user-inputted language. We contribute a novel framework, SplaTraj, which formulates the generation of images within photorealistic environment representations as a continuous-time trajectory optimization problem. Costs are designed so that a camera following the trajectory poses will smoothly traverse through the environment and render the specified spatial information in a photogenic manner. This is achieved by querying a photorealistic representation with language embedding to isolate regions that correspond to the user-specified inputs. These regions are then projected to the camera's view as it moves over time and a cost is constructed. We can then apply gradient-based optimization and differentiate through the rendering to optimize the trajectory for the defined cost. The resulting trajectory moves to photogenically view each of the specified objects. We empirically evaluate our approach on a suite of environments and instructions, and demonstrate the quality of generated image sequences.
\end{abstract}

%%%%%%%%%%%%%%%%%%%%%%%%%%%%%%%%%%%%%%%%%%%%%%%%%%%%%%

\section{INTRODUCTION}
\label{sec:introduction}
Autonomous agents operating in unknown environments need to construct internal representations of their operating environment. Traditionally, these representations, such as occupancy or semantic maps, can be difficult for the untrained eye to interpret. Recent advances in computer vision have led to the development of \emph{photorealistic} 3D environment representations \cite{mildenhall2020nerf}, which enable high-fidelity visualizations of the robot's surroundings. Of these representations, Gaussian Splatting models \cite{kerbl3Dgaussians} have emerged as the model of choice due to their ability to render photorealistic images efficiently. There is great excitement within the community to advance methodologies for constructing Gaussian Splatting models \cite{zhang2024darkgs} and embedding additional properties into them. Yet many unsolved challenges exist around how best to leverage these representations for downstream tasks. 

A valuable downstream task for Gaussian Splatting is to take advantage of its visual realism to extract visually accurate sequences of images. This paper tackles this challenge. Specifically, we study the problem of generating a sequence of photorealistic images from a Gaussian Splatting model that smoothly displays a sequence of objects that human users can semantically specify. In particular, we introduce the SplaTraj framework, which formulates a trajectory optimization problem that attempts to optimally position the camera, such that it smoothly moves through the reconstructed environment while accurately pointing, in order, to each of the semantically specified objects within the scene. 

 \begin{figure}[t]
    \centering
    \includegraphics[width=\linewidth]{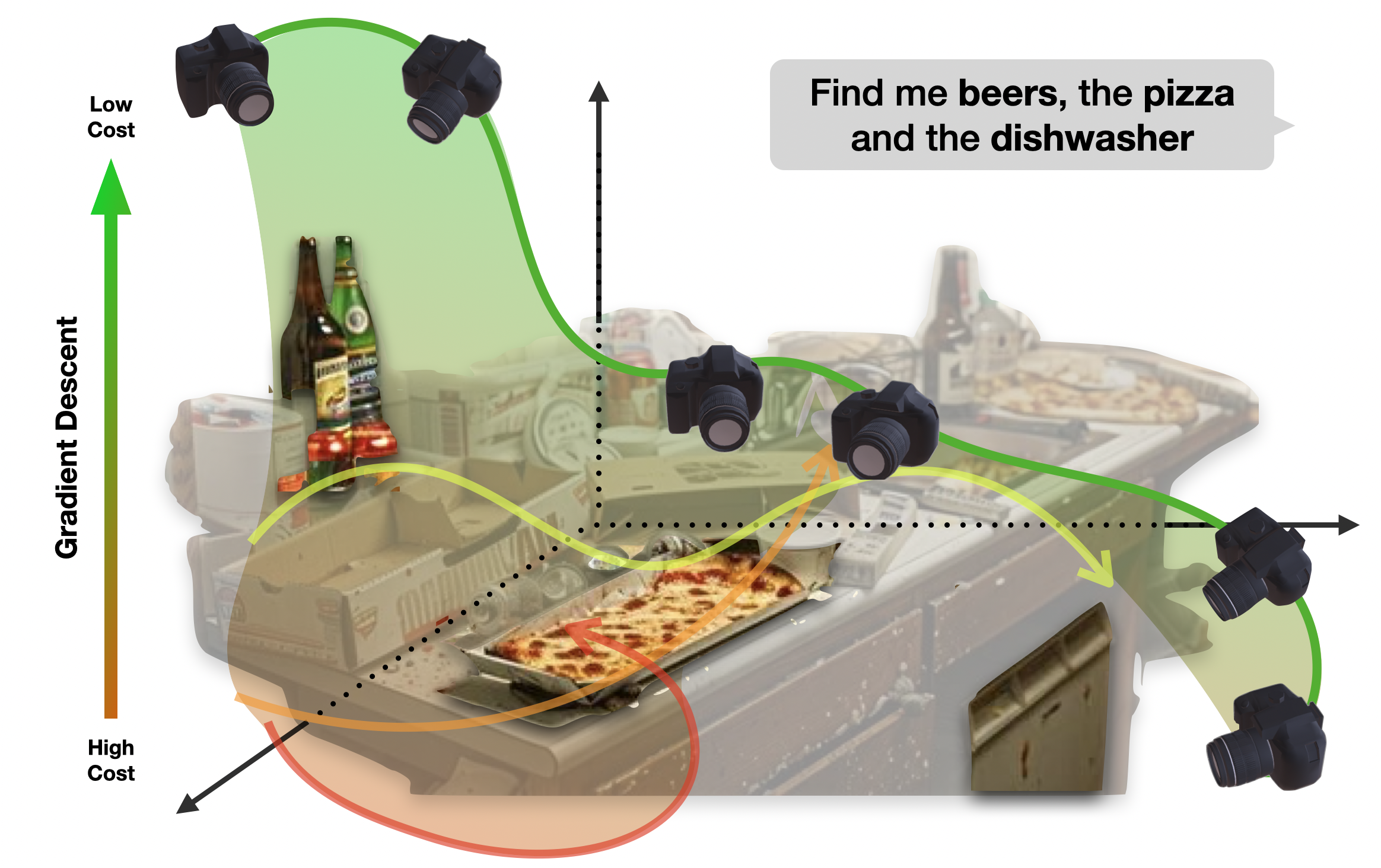}
    \caption{We propose SplaTraj, which enables users to provide semantic instructions consisting of a sequence of objects. This is then formulated as a gradient-based trajectory optimization problem within a photorealistic Gaussian Splatting model, whose solution smoothly shows us each of the objects.}
    \label{fig:splatraj_intro}
\end{figure}

% mention that a occlusion-avoiding design can lead to finer construction of the representation, which can serve better in active learning context.
SplaTraj operates on top of  Gaussian Splatting models that encode visual language features into the environment reconstruction. Then, embedded semantics in the environment are compared with user-inputted language semantics to identify objects and regions within the environment that are relevant to the user's intended instruction. These identified objects are rendered into the view of the camera and carefully designed costs evaluate the quality of their visual placement. SplaTraj parameterized camera trajectories as continuous time-varying functions. By applying gradient-based optimizers to the cost-optimal motion trajectory, we can differentiate through both the rendering equation and the trajectory parameterization. This enables us to obtain a model that describes the optimal camera motion through the environment so that we can photogenically capture all of the objects specified by the user. A simple overview is illustrated in \cref{fig:splatraj_intro}.

Concretely, our technical contributions are:
\begin{itemize}
    \item The SplaTraj framework, which formulates generating image sequences as a continuous trajectory optimization problem over camera poses;
    \item Methodology to extract environment structure from user-inputted semantics and then incorporate the structures as an optimization cost for trajectory optimization;
    \item Empirical evaluation of SplaTraj to generate images in a variety of benchmark Gaussian Splatting environments.
\end{itemize}

\section{RELATED WORK}
\label{sec:relatedwork}
\textbf{Robot Representations and Photorealistic Representations:} Robots operating in unknown environments rely on constructing internal representations of the environment. These have generally been representations of occupancy \cite{OccupancyGridMaps, HM, wright2024vprism}, surfaces \cite{SDF}, or motion patterns \cite{sptemp, KTM}. However, such representations are often difficult for the untrained eye to interpret. This has led to photorealistic representations, such as Neural Radiance Fields (NeRFs) \cite{mildenhall2020nerf}, and more 3D Gaussian Splatting \cite{kerbl3Dgaussians}. Subsequent works \cite{mueller2022instant,zhang2024darkgs,zhang2024recgs} in this area augment the capabilities of photorealistic reconstructions. Our work leverages efforts in embedding language features \cite{radford2021learning} into realistic representations \cite{qin2024langsplat3dlanguagegaussian}. 

\textbf{Camera Optimization:} Our work is tangentially related to camera placement optimization, which has been an active area of research. \cite{fleishman1999automatic} propose an automatic camera placement method for generating image-based models from scenes with known geometry. Albahri et. al \cite{albahri2017simulation} explore camera placement within buildings, focusing on maximizing. Another work in this space, \cite{bodor2005multi} addresses the optimal placement of cameras for human activity recognition, maximizing the observability. Our work differs from this body of work in that our optimization occurs within a photorealistic reconstruction and is driven by user-provided semantics.

\textbf{Trajectory Optimization:} Trajectory optimization is commonly used within motion planning \cite{Motion_planning, ompl, pdmp} and control to obtain motion sequences to reach a certain goal. Prominent trajectory optimization methods within motion planning, including TrajOpt \cite{Schulman2013FindingLO}, CHOMP \cite{CHOMP}, STOMP \cite{Kalakrishnan2011STOMPST}. Trajectory optimization for a finite horizon also appears within model predictive control \cite{NMPC}. Prominent trajectory optimization methods include MPPI \cite{MPPI}. Recent approaches have also been introduced \cite{RMPs, geoFabs, GeoFab_gloabL_opt, Fast_diff_int} to generate motions reactively. This shortens the time horizon even further. Classic trajectory optimization formulates an objective that finds a minimal distance path. Our work differs from these approaches in that we formulate a cost within trajectory optimization that optimizes object placement within the camera view, by differentiating the camera rendering.

\section{PRELIMINARIES}
\label{sec:prelim}

\subsection{Open Querying on Images}
Open query refers to the ability of the model to handle
arbitrary text inputs for image retrieval or classification tasks, without being limited to a predefined set of categories or
labels. Contrastive Language-Image Pre-Training \cite{clip} is commonly used for tasks of this class.
CLIP jointly trains an image encoder and a text encoder to predict the correct pairings of a batch of training examples. CLIP is typically used as a plug-and-play encoder. It is trained on a large-scale dataset of 400 million image-text pairs, enabling it to learn visual concepts from natural language descriptions. In this paper, we also use the pre-trained encoder provided in CLIP to generate embeddings from semantic inputs.

\subsection{Gaussian Splatting and Language Field}
Gaussian Splatting\cite{kerbl3Dgaussians} explicitly represents a 3D scene as  a  collection  of  anisotropic  3D  Gaussians,  with  each Gaussian $G(x)$ characterized by a mean $\mu \in \mathcal{R}^3$ and a co-variance matrix $\Sigma$:
\[
G(x)=\exp \left(-\frac{1}{2}(x-\mu)^{\top} \Sigma^{-1}(x-\mu)\right)
\]
For each camera pose in the image sequence, an image can be synthesized from the following rendering equation \cite{kerbl3Dgaussians}:
\begin{equation}
\label{eq:rendereq}
    \hat{I} = \sum_{\substack{i\in\mathcal{N}}}c^i\alpha^i \overset{i-1}{\prod_{\substack{j=1}}} (1-\alpha^j)
\end{equation}

where $c^i$ is the color of the $i$-th Gaussian, $\mathcal{N}$ denotes the Gaussians in the tile, 
% $C(v)$ is the rendered color at pixel $v$,
and $\alpha^i = o^i G^i_{2 D}(v)$. Here $o^i$ is the opacity of the $i$ th Gaussian and $G^i_{2 D}(\cdot)$ represents the function of the $i$-th Gaussian projected onto 2D. 
Recent research has enhanced 3D Gaussians with language features \cite{zheng2024gaussiangrasper3dlanguagegaussian}\cite{qin2024langsplat3dlanguagegaussian}. 
$$
\hat{I_l} =\sum_{i\in\mathcal{N}} l^i \alpha^i \prod_{j=1}^{i-1}\left(1-\alpha^j\right)
$$

where $l^i$ is the open-vocabulary feature embedding of i-th 3D gaussian primitives and $\hat{I_l}$ represents the rendered open-vocabulary feature embedding at pixel $u$. The intuition here is simple: positions in the scene are mapped to language features rather than colors. We use upper script index notation specifically for the 3D Gaussian index, to distinguish it from other index notations in the paper.

% Unlike traditional methods that directly learn CLIP embeddings, LangSplat introduces a novel two-step process. Initially, it employs a scene-wise language autoencoder, which compresses the scene information into a latent space. Subsequently, language features are learned within this scene-specific latent space. This strategy effectively alleviates the substantial memory demands typically associated with explicit modeling of language features.

%  In LangSplat\cite{qin2024langsplat3dlanguagegaussian}, M. Qin et al. use the segmentation and semantic embedding of SAM -- foundation model for image segmentation -- to train an OpenClip to learn language fields. Instead of directly learning CLIP embeddings, LangSplat first trains a scene-wise language autoencoder and then learns language features on the scene-specific latent space, thereby alleviating substantial memory demands imposed by explicit modeling.
% LangSplat can accurately capture object boundaries and provide precise 3D language fields in the rendered language feature image.  

\section{METHODOLOGY}
\label{sec:methodology}
\begin{figure*}
    \centering
    \includegraphics[width=0.88\linewidth]{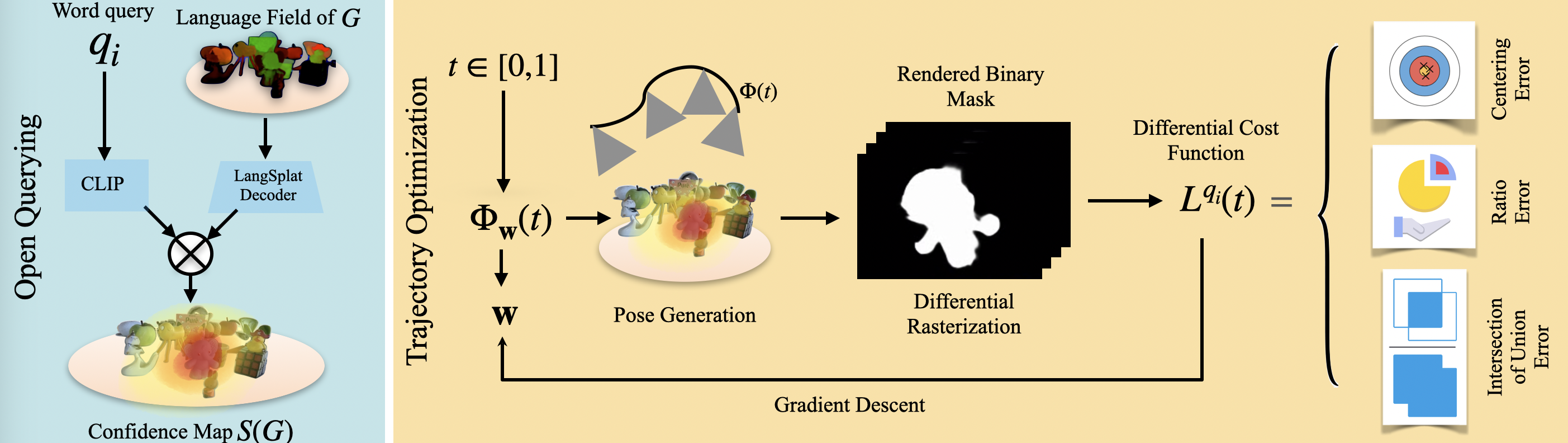} 
    \caption{Pipeline overview: Our method consists of two main steps, vocabulary querying and gradient-based trajectory optimization. We generate a 3D object mask based on language embeddings and trained language fields in the vocabulary query step. In trajectory optimization, we use the differential renderer to generate rendering-based cost and gradient descent to update the trajectory coefficients. The 3D confidence heatmap derived in preprocessing is fixed in the downstream differential rasterization. }
    \label{fig:pipline}
\end{figure*}

In this section, we proposed our differentiable optimization method to solve photogenic trajectory generation of the open-queried objects represented in 3D Gaussians. The overall framework is shown in Fig. \ref{fig:pipline}. 

%\section{Method} % Change the section title to the name of the method

\subsection{Problem Formulation}

This paper addresses the optimization problem of generating realistic camera trajectories in environments represented by language-annotated 3D Gaussians \cite{qin2024langsplat3dlanguagegaussian}. Given a user-specified query of various modalities such as text, \( Q = (q_1, q_2, \ldots, q_n) \), our goal is to determine a trajectory of camera pose \( \mathbf{\Phi} \in \mathrm{SE(3)}  
% = [r_x (t), r_y(t), r_z(t), x(t),y(t),z(t)]
\)  over a normalized period $t \in [0,1]$.

The camera poses should sequentially focus on the objects corresponding to each \( w_i \) within the specific time intervals \( T_i = \left[ \frac{i-1}{n}, \frac{i}{n} \right] \) ($\bigcup_{i} T_i = 1, \bigcap_{i} T_i = \emptyset$). Our objective is to render photogenic videos during these intervals.

In this context, a photo is considered more photogenic if the specified object is more centered in the image and occupies a portion closer to a user-defined ratio. Formally, we seek to optimize the camera trajectory such that each object \( w_i \) is appropriately captured in its designated time slot, resulting in a series of well-composed images.

The challenge lies in dynamically adjusting the camera trajectory to meet these criteria while smoothly transitioning between different objects in the sequence.
% representation of language-annotated 3D Gaussians, prompt words of target objects ...
% the top-down optimization problem formulation goes here

\subsection{Semantic Map Extraction}% or open vocabulary query 
We seek to ground semantic specifications given by the user to physical coordinates within our splatting model. We leverage the semantic information encoded in the 3D Gaussians to generate a differentiable object mask for each word query, following a similar procedure in \cite{kerr2023lerflanguageembeddedradiance}. %However, since the decoder is inherently invariant towards the order and structure of the language feature, we directly decode pre-trained compressed 3D language features of each Gaussian $G^j$, recovering the language embedding $\varpi_{G}^j$.

 The relevancy score on a Gaussian $G^j$ is defined as a softmax score on the 
 \begin{equation}
 S(G^j) = \min_{i}\frac{\exp (\varpi_{G}^j\cdot \varpi_{qry})}{\exp(\varpi_{G}^j\cdot \varpi_{qry})+\exp(\varpi_{G}^j\cdot \varpi^i_{canon})}, 
\end{equation}
where $\varpi^i_{canon}$ is the  CLIP  embeddings of a predefined canonical phrase chosen from “object”, “things”, “stuff”, and “texture”. Experimentally, we found that a DBScan algorithm \cite{ester1996density} on top relevancy score percentile Gaussians can filter out irrelevant noises in 3D space. The relevancy score suffered from the score scale ambiguity and spread outliers in 3D space, which would cause unclear mask edges and outliers in the rendering result, leading to downstream failure.  A binary channel for object prompt $q_i$ based on the original 3D Gaussian field is attained in the filtering procedure, where the $j^{\text{th}}$ entry of the $q_i$ binary channel is $b^j_i = \mathbb{I}_{G^j \in \mathcal{N}_i}$, where $\mathbb{I}_{\{A\}}$ is the indicator of event A, $\mathcal{N}_i$ is the set of Gaussians after filtering of prompt $q_i$. 

 We rendered 2D binary mask via the following rendering equation:
 \begin{equation}
I_b =\sum_{i\in\mathcal{N}} b^i \alpha^i \prod_{j=1}^{i-1}\left(1-\alpha^j\right).\label{eqn:render}
\end{equation}
 By rendering the mask onto the camera view, we can reason about properties in the camera's view. This includes the ratio of the object within the frame, giving an indication of how photogenic the rendered image is.

% \begin{figure}[!ht]
%     \centering
%     \subfigure[]{%
%         \includegraphics[width=0.15\textwidth]{example-image}
%     }%
%     \hfill
%     \subfigure[w/o filtering]{%
%         \includegraphics[width=0.15\textwidth]{example-image}%
%     }%
%     \hfill
%     \subfigure[w/ filtering]{%
%         \includegraphics[width=0.15\textwidth]{example-image}%
%     }%
%     \caption{Rendered image, rendered object mask without DBscan filtering, rendered object mask with DBscan filtering  }
%     \label{fig:images}
% \end{figure}

\subsection{Continuous Trajectory Representation}
 Many trajectory optimization problems express trajectories as a sequence of waypoints. This typically requires \emph{a priori} discretization of the trajectory at some fixed time resolution. Here, we take an alternative approach and represent the motion trajectory of the camera as a continuous function, mapping from a normalized time parameter to the camera pose. This enables us to generate trajectories of any desired resolution. 
 
 To represent a camera trajectory vector function in 6 dimensions, we use 6 independent functions to model each variable of
\begin{equation}
\mathbf{\Phi}(t) = [r_x (t), r_y(t), r_z(t), x(t),y(t),z(t)]^{\top}, 
\end{equation} 
where $r_x, r_y, r_z$ are rotation vectors of transformation from the world coordinate to the camera coordinate, $x,y,z$ are translation components of \( t \in [0, 1] \). We represent this function concisely as a combination of Squared Exponential Radial Basis Functions (RBFs). Specifically, the j-th entry of $\mathbf{\Phi}(t)$ could be represented by a weighted sum of RBFs: \begin{equation}
[\mathbf{\Phi}(t)]_j 
=\sum_{i=1}^N w_i^j \psi_i\left(t\right) 
= \mathbf{w}_j^{\top} \mathbf{\Psi}(t),
\end{equation}
where $\mathbf{w}_j = (w_1^j, w_2^j, ..., w_N^j) $ is the weight vector associated with the j-th entry, and $\mathbf{\Psi(t)} = (\psi_1(t), \psi_2(t), ..., \psi_N(t)) $ is the packed RBFs, assigned with entry-specific weight vector $\mathbf{w}_j$. 
We use the Gaussian function as our basis function as
\begin{equation}
\psi_i\left(t\right) = \exp\left(- \left(  \frac{(t - t_i)^2 }{2 \sigma^2} \right) \right)\label{eqn:traj_form} ,
\end{equation}
where $t_i = \frac{i}{n}-\frac{1}{2}, i = 1,2, \ldots, n$ is the center of the time interval $T_i$, hyper-parameter $\sigma$ is the standard deviation of the Gaussian distribution.
This representation offers a smooth and flexible trajectory generation by interpolating between control points with a smooth basis function. Notably, our trajectory formulation enables us to query and obtain a camera pose at arbitrary time and at arbitrary time resolution. 

\subsection{Cost Function}
We constructed an optimization problem to generate trajectories that minimize the defined cost so that the camera can sequentially fixate on objects corresponding to language inputs given by the user \( Q = (q_1, q_2, \ldots, q_n) \), we are optimizing the following cost function, defined as a piece-wise function distributing each object prompt to the time interval $T_i = \left[ \frac{i-1}{n}, \frac{i}{n} \right]$, specifically,
\begin{equation}
        \arg \min_{\mathbf{w}} \quad \sum_{i=1}^{n} \int_{T_i} L^{q_i}(\mathbf{\Phi}_{\mathbf{w}}(t)) dt, 
\end{equation}
where 
% $\mathbf{\Phi}_{\mathbf{w}} (t)$ is the trajectory represented as a vector value function of normalized time $t \in [0,1]$, and $\mathbf{w}$ is the parameter of the trajectory representation $\phi(t)$, given in \cref{eqn:traj_form}. 
$L^{q_i}(\cdot)$ is the cost for prompt word query $q_i$. To obtain the cost, we shall render the identified semantically-relevant 3D structure to the camera's view to obtain a binary mask. We denote the mask as $I_b$ and compute by evaluating \cref{eqn:render}.

We define $L^{q_i}$ as the sum of multiple cost terms,
\begin{equation}
L^{q_i}=L_{TCE}+L_{TRE}+L_{upright}+\alpha L_{prior}. 
\end{equation}
Here, $\alpha$ is a weight of the cost term $L_{prior}$ which gradually decays. The definitions of each cost term are elaborated below.
% Note that we define our cost function as dependent only on the pose vector. In practice, we penalize pose-wise cost from a sampled batch of poses from the camera trajectory and calculate the error. 

% We designed rendering-based cost, which was based on the rendered 2D images, and 3D-based cost, which was based on 3D information that was primarily used in 3D trajectory planning.  We argue that the 2D image cost could effectively guide the camera to a photogenic position by differentiating through the rendering equation. 

% \begin{equation}
% L^{q_i} = L_{render}^{q_i}  + L_{3D}^{q_i} 
% \end{equation}

The rendering-based cost includes the center cost and the ratio cost. Here we render of object that we identify as relevant to the user's prompt as a binary mask, denoted $I_b$ and computed by evaluating \cref{eqn:render}. Then $L^{q_i}$ is the sum of the following terms: 
\subsubsection{Target Centralizing Error (TCE) Cost}
\begin{equation}
L_{TCE} (I_b) = || (\frac{c_x}{H}, \frac{c_y}{W}) - (\frac{1}{2}, \frac{1}{2})||_2,
\end{equation}
where $(c_x, c_y) = center(I_b)$ are pixel coordinates of the center of the mask. 
\subsubsection{Target Ratio Error (TRE) Cost}
\begin{equation}
L_{TRE} (I_b) = || \frac{sum(I_b)}{HW} - r_t||_2,
\end{equation}
where $sum(I_b)$ counts the area of the 2D binary mask $I_b$.
% The 3D-based cost includes the uprightness cost and the obstacle-avoiding cost.
\subsubsection{Uprightness Cost} The camera's orientation can be regulated to an upright position by the following cost term
\begin{equation}
L_{upright} = -<\mathbf{e_z}, \mathbf{R}  \mathbf{e_x}>,
\end{equation}
Where we use the inner product between the world unit upward direction axis $e_z$ and the camera's unit upright axis $\mathbf{R} \mathbf{e_x}$ where $\mathbf{e}$ are unit coordinate vector and $\mathbf{R}$ denotes the rotation from camera frame to world frame.
% \subsubsection{Obstacle Avoiding Cost}
% \[
% L_{obs}= \frac{\sum \left( e^{\bar{s}_{i}} \cdot \frac{1}{d_i} \right)}{\sum e^{\bar{s}_{i}} + \epsilon}
% \]
% where:
% \[\mathbf{d}_i = |\mathbf{p}_i - \mathbf{c}| \quad \text{if} \quad |\mathbf{p}_i - \mathbf{c}| < r \]

% \[\bar{s}_i = \text{mean}(\text{obs_sc}[i]) \quad \text{for all valid} \quad i  \]
% \[\mathbf{p}_i = \text{obs_xyzs}[i]\]
% \[\mathbf{c} = \text{cam.camera_center} \]

\subsubsection{Diminishing Prior Cost} We warm-start the optimization via a prior cost which guides the camera toward the physical adjacency of the target objects. This is given:

\begin{equation}
L_{prior} = -\max(r, \|\mathbf{c} - \mathbf{o}\|_2) - \frac{(\mathbf{c} - \mathbf{o})^{C}_z}{\|\mathbf{c} - \mathbf{o}\|_2}.
\end{equation}
The first term in the $L_{prior}$ calculates a cost based on the proximity of the camera to the object, with a minimum distance threshold defined by the radius; The second term of the $L_{prior}$ calculates the angle deviation from the object, specifically measured by the inner product between camera direction unit vector with the normalized vector pointing from object to the camera center.  

% \subsection{Optimization Procedure}
 During optimization, we utilize the differential renderer to calculate the rendering-based cost $L_{TCE}$, $L_{TRE}$ and $L_{IoU}$, whose gradient with respect to trajectory coefficient $\frac{\partial L}{\partial \mathbf{w}}$ can be attained from the differential renderer. We warm-start the progress by giving an exponentially decaying coefficient to differentiable prior term $L_{prior}$. The total cost gradient with respect to trajectory coefficient $\mathbf{w}$ would update the trajectory coefficient via Adam optimizer\cite{kingma2017adammethodstochasticoptimization}, iteratively refining the camera trajectory.

\section{EMPIRICAL RESULTS}
\label{sec:experiments}

\subsection{Experiments Overview}
In this section, we seek to investigate empirically:
\begin{itemize}
    \item Whether the image cost design leads to object-centered, properly distanced, and occlusion-avoiding rendering images; (\Cref{subsec:single_pose}) 
    \item If the photogenic objective of the optimization would lead to new challenges and properties, such as multiple optimal solutions;
    \item Whether the trajectory representation used is amenable to optimization with respect to the objective defined.% need to rephrase
    % \item Whether the generated trajectories are smooth and collision-avoiding.
\end{itemize}
We validate our method in both single-pose optimization settings in Section \ref{subsec:single_pose} and trajectory optimization settings in Section \ref{subsec:traj_opt}. 
%

% \subsection{Practical Implementation}
% check device specifications. 

% we can talk about 
\subsection{Single Pose Optimization}
\label{subsec:single_pose}
To ensure that the pose-wise cost function components are correct, we first design a single pose optimization to validate our method. Instead of optimizing the trajectory weight coefficient, we optimize on individual poses, using the same cost function defined in section \ref{sec:methodology}.  The objective function is therefore reduced to
$ \arg \min_{\mathbf{\Phi}}  L^{q_i}(\mathbf{\Phi})$, where $  \mathbf{\Phi} = [r_x, r_y, r_z, x,y,z] $ is the optimization variable.

With each selected query prompt $q_i$, we preprocess the scene dataset by extracting the semantic mask. During optimization, we calculate the cost value and gradient of the current pose and run gradient descent on the cost manifold $L^{q_i}$. We use the ADAM optimizer to evaluate the optimization result's TCE, TRE, and IoU score after 400 iterations. We also did an ablation study to show the effect of each optimization cost term, as shown in table \ref{tab:single_cam}. We first show theimpact of 3D prior terms, which includes all terms derived from 3D prior information without decaying weights; We then apply TCE and TRE regulating terms separately and jointly, with 3D prior terms with diminishing weight through steps.  Our results show:
\begin{table}
    \centering
    \parbox{\linewidth}{
    \centering
    \begin{tabular}{c|c|c|c|c}
         & \textbf{Scenarios} & $\overline{TCE}$ $\downarrow$ & $\overline{TRE}$ $\downarrow$ & $\overline{IoU}$ $\uparrow$\\ \hline
         \multirow{2}{*}{\textbf{3D Prior}} & Teatime & $0.066 $ & $0.321 $  & $0.531$ \\
         & Figurines & $0.051 $ & $0.220$ & $0.522 $\\
         & Kitchen & $0.151 $ & $0.262 $ & $0.527$\\\hline
         \multirow{2}{*}{\textbf{TCE}} & Teatime & $\mathbf{0.014} $ & $ 0.119 $ & $ 0.805 $\\
         & Figurines & $\mathbf{0.010} $ & $0.117 $ & $0.480 $\\
         & Kitchen & $0.120 $ & $0.112 $ & $0.617 $\\\hline
         \multirow{2}{*}{\textbf{TRE}} & Teatime & $0.018 $ & $0.068 $ & $ 0.808$\\
         & Figurines & $0.115$ & $0.126$ & $0.384$\\
         & Kitchen & $\mathbf{0.021}$ & $\mathbf{0.057}$ & $0.711$\\\hline
         % \multirow{2}{*}{\textbf{IoU}} & Teatime & $0.100 \pm 0.162 $ & $0.056 \pm 0.046$ & $0.823 \pm 0.176$\\
         % & Figurines & $0.219 \pm 0.375$ & $0.124 \pm 0.019$ & $0.672 \pm 0.312$\\
         % & Waldo Kitchen & $0.268 \pm 0.392$ & $0.076 \pm 0.053$ & $0.785 \pm 0.117$\\\hline
         \multirow{2}{*}{\textbf{TCE+TRE}} & Teatime & $0.029 $ & $\mathbf{0.057}$ & $\mathbf{0.867}$ \\
         & Figurines & $0.081$ & $\mathbf{0.107 }$ & $\mathbf{0.760}$\\
         & Kitchen & $0.098$ & $0.065$ & $\mathbf{0.751}$\\\hline
         % \multirow{2}{*}{\textbf{TCE+IoU}} & Teatime & $0.060 \pm 0.123 $ & $0.117 \pm 0.072$ & $0.888 \pm 0.044$ \\
         % & Figurines & $0.042 \pm 0.037$ & $0.117 \pm 0.018$ & $0.650 \pm 0.284$\\
         % & Waldo Kitchen & $0.016 \pm 0.009$ & $0.091 \pm 0.048$ & $0.781 \pm 0.095$\\\hline
         % \multirow{2}{*}{\textbf{TRE+IoU}} & Teatime & $0.101 \pm 0.250 $ & $0.087 \pm 0.081$ & $0.908 \pm 0.069$ \\
         % & Figurines & $0.209 \pm 0.361$ & $0.095 \pm 0.047$ & $0.715 \pm 0.312$\\
         % & Waldo Kitchen & $0.089 \pm 0.205$ & $0.057 \pm 0.032$ & $0.816 \pm 0.050$\\\hline
         
    \end{tabular}}
    \caption{Result from single camera pose optimization with randomly selected queries, 10 repetitions with randomized initial decision variable. }
    \label{tab:single_cam}
\end{table}

\begin{figure*}[t]
    \centering
    \setlength{\tabcolsep}{1pt} % Reduce space between subfigures
    \renewcommand{\arraystretch}{0} % Reduce space between rows
    
    \begin{tabular}{cc|ccccccccc}
        % & &  $\mathbf{t = \frac{1}{18}}$  & $\mathbf{t = \frac{1}{6}}$  & $\mathbf{t = \frac{5}{18}}$ & $\mathbf{t = \frac{7}{18}}$  & $\mathbf{t = \frac{1}{2}}$  & $\mathbf{t = \frac{11}{18}}$ & $\mathbf{t = \frac{13}{18}}$  & $\mathbf{t = \frac{5}{6}}$  & $\mathbf{t = \frac{17}{18}}$ \\ \hline
        & &  $\mathbf{t = 0.06}$  & $\mathbf{t = 0.17}$  & $\mathbf{t = 0.28 }$ & $\mathbf{t = 0.39}$  & $\mathbf{t = 0.5}$  & $\mathbf{t = 0.61}$ & $\mathbf{t = 0.72}$  & $\mathbf{t = 0.83}$  & $\mathbf{t = 0.94}$ \\ 
        \vspace{0.1cm}\\
        \hline
        \multirow{3}{*}{\adjustbox{angle=90}{\textbf{\quad \quad Teatime}}} & & 
        \includegraphics[width=0.1\linewidth]{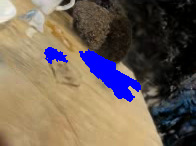} &
        \includegraphics[width=0.1\linewidth]{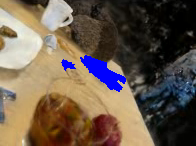} &
        \includegraphics[width=0.1\linewidth]{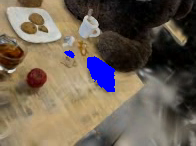} &
        \includegraphics[width=0.1\linewidth]{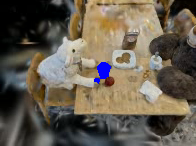} &
        \includegraphics[width=0.1\linewidth]{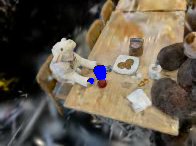} &
        \includegraphics[width=0.1\linewidth]{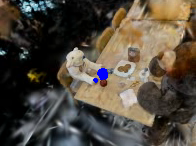} &
        \includegraphics[width=0.1\linewidth]{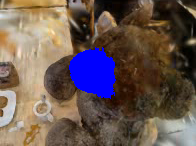} &
        \includegraphics[width=0.1\linewidth]{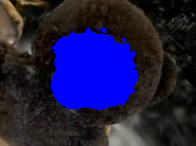} &
        \includegraphics[width=0.1\linewidth]{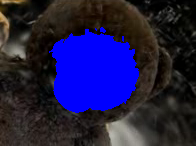} 
        \\
        & &\includegraphics[width=0.1\linewidth]{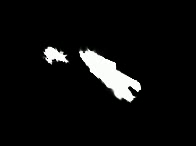} &
        \includegraphics[width=0.1\linewidth]{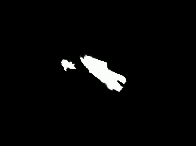} &
        \includegraphics[width=0.1\linewidth]{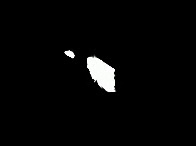} &
        \includegraphics[width=0.1\linewidth]{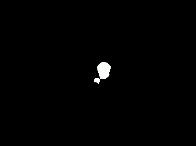} &
        \includegraphics[width=0.1\linewidth]{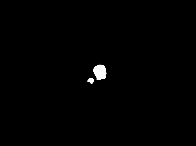} &
        \includegraphics[width=0.1\linewidth]{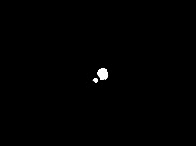} &
        \includegraphics[width=0.1\linewidth]{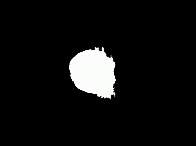} &
        \includegraphics[width=0.1\linewidth]{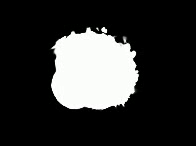} &
        \includegraphics[width=0.1\linewidth]{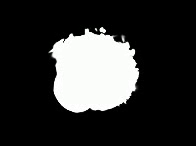} \\ 
        % \vspace{0.15cm}\\
         \text{\quad} & & \multicolumn{3}{c|}{``napkins''} &\multicolumn{3}{|c|}{``glass of water''} &
         \multicolumn{3}{|c}{``bear nose''}  \\ 
         % \vspace{0.15cm}\\
        \hline
        \multirow{3}{*}{\adjustbox{angle=90}{ \textbf{\quad \quad Kitchen}}} & &
        \includegraphics[width=0.1\linewidth]{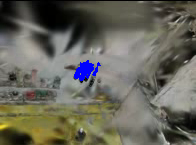} &
        \includegraphics[width=0.1\linewidth]{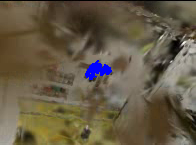} &
        \includegraphics[width=0.1\linewidth]{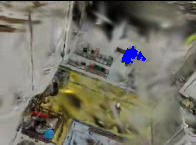} &
        \includegraphics[width=0.1\linewidth]{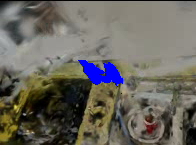} &
        \includegraphics[width=0.1\linewidth]{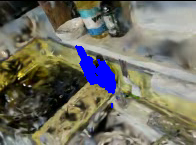} &
        \includegraphics[width=0.1\linewidth]{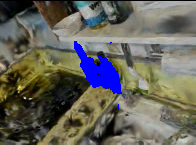} &
        \includegraphics[width=0.1\linewidth]{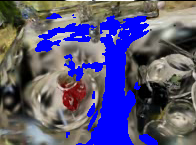} &
        \includegraphics[width=0.1\linewidth]{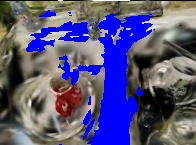} &
        \includegraphics[width=0.1\linewidth]{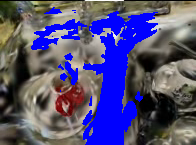} 
        \\
        & &\includegraphics[width=0.1\linewidth]{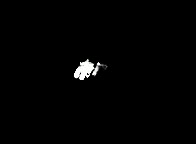} &
        \includegraphics[width=0.1\linewidth]{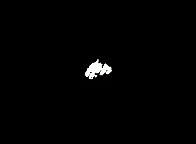} &
        \includegraphics[width=0.1\linewidth]{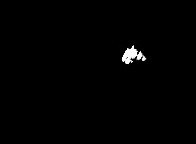} &
        \includegraphics[width=0.1\linewidth]{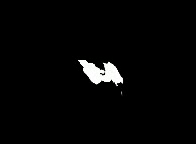} &
        \includegraphics[width=0.1\linewidth]{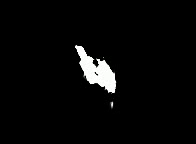} &
        \includegraphics[width=0.1\linewidth]{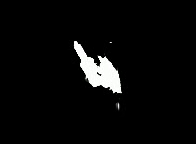} &
        \includegraphics[width=0.1\linewidth]{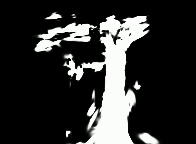} &
        \includegraphics[width=0.1\linewidth]{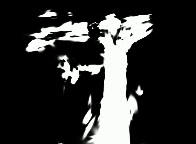} &
        \includegraphics[width=0.1\linewidth]{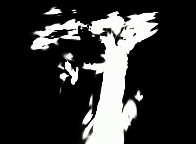} \\ 
         
        % & & & \text{``knives''} & & &\text{``oil bottles''} & & &\text{``sink''} & \\ 
        \text{\quad} & & \multicolumn{3}{c|}{``knives''} &\multicolumn{3}{|c|}{``oil bottles''} &
         \multicolumn{3}{|c}{``sink''}  \\ 
         \vspace{0.05cm}\\
        \hline
        \multirow{3}{*}{\adjustbox{angle=90}{\textbf{\quad \quad Figurines}}} & & 
        \includegraphics[width=0.1\linewidth]{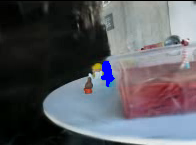} &
        \includegraphics[width=0.1\linewidth]{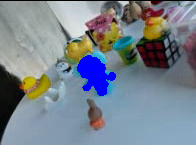} &
        \includegraphics[width=0.1\linewidth]{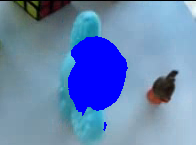} &
        \includegraphics[width=0.1\linewidth]{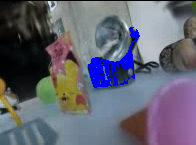} &
        \includegraphics[width=0.1\linewidth]{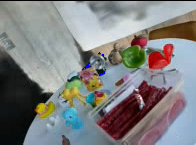} &
        \includegraphics[width=0.1\linewidth]{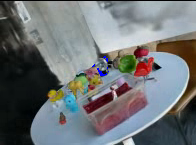} &
        \includegraphics[width=0.1\linewidth]{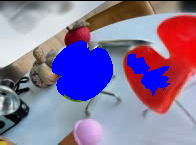} &
        \includegraphics[width=0.1\linewidth]{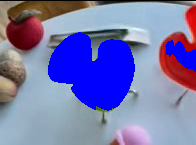} &
        \includegraphics[width=0.1\linewidth]{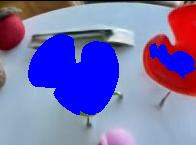} 
        \\
        & &\includegraphics[width=0.1\linewidth]{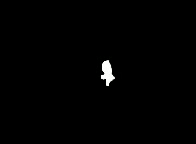} &
        \includegraphics[width=0.1\linewidth]{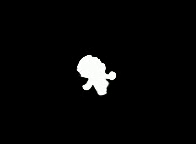} &
        \includegraphics[width=0.1\linewidth]{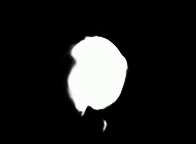} &
        \includegraphics[width=0.1\linewidth]{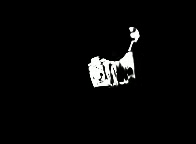} &
        \includegraphics[width=0.1\linewidth]{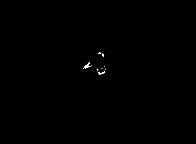} &
        \includegraphics[width=0.1\linewidth]{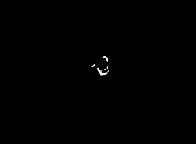} &
        \includegraphics[width=0.1\linewidth]{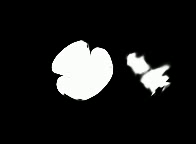} &
        \includegraphics[width=0.1\linewidth]{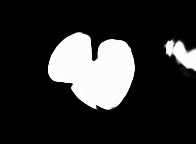} &
        \includegraphics[width=0.1\linewidth]{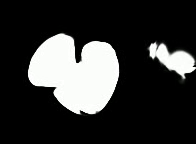} \\
        %  \vspace{0.15cm}
        % & & & \text{``toy elephant''} & & &\text{``old camera''} & & &\text{``chair''} & \\
        \text{\quad} & & \multicolumn{3}{c|}{``toy elephant''} &\multicolumn{3}{|c|}{``old camera''} &
         \multicolumn{3}{|c}{``chair''}  \\ 
        
        % \hline
    \end{tabular}

    \caption{Visualization of SplaTraj results across Scenarios: We selected 9 evenly spaced keyframes from the trajectory sequence for visualization. We present the rendered masks and rendered RGB images with overlaid blue masks. We observe that as the camera moves along the resulting trajectory, each of the semantically specified objects appear prominently in view.}
    \label{fig:subfigures}
\end{figure*}

% think about argument around this experiment
\begin{itemize}
    \item TCE cost term applied in the optimization can centralize objects better, and enabling centering cost or ratio regulation cost might improve the performance of the corresponding aspect.
    \item The benefit of TCE and TRE can be additive, and combining the two regulating terms generally finds the best pose. 
    \item The IoU alone does not significantly improve the overall performance. In the experiment, we observe the non-smoothness of the IoU cost function, leading to huge steps in gradient descent that lead out of the low-cost region, similar issues have been discussed in the object detection field \cite{arif2023directlyoptimizingioubounding}.
\end{itemize}

\subsubsection{Stochastic Gradient Langevin Dynamics (SGLD) \cite{SGLD}}
\begin{figure}[t]
    \centering
    \subfigure[Initialized Pose]{%
        \includegraphics[width=0.48\linewidth]{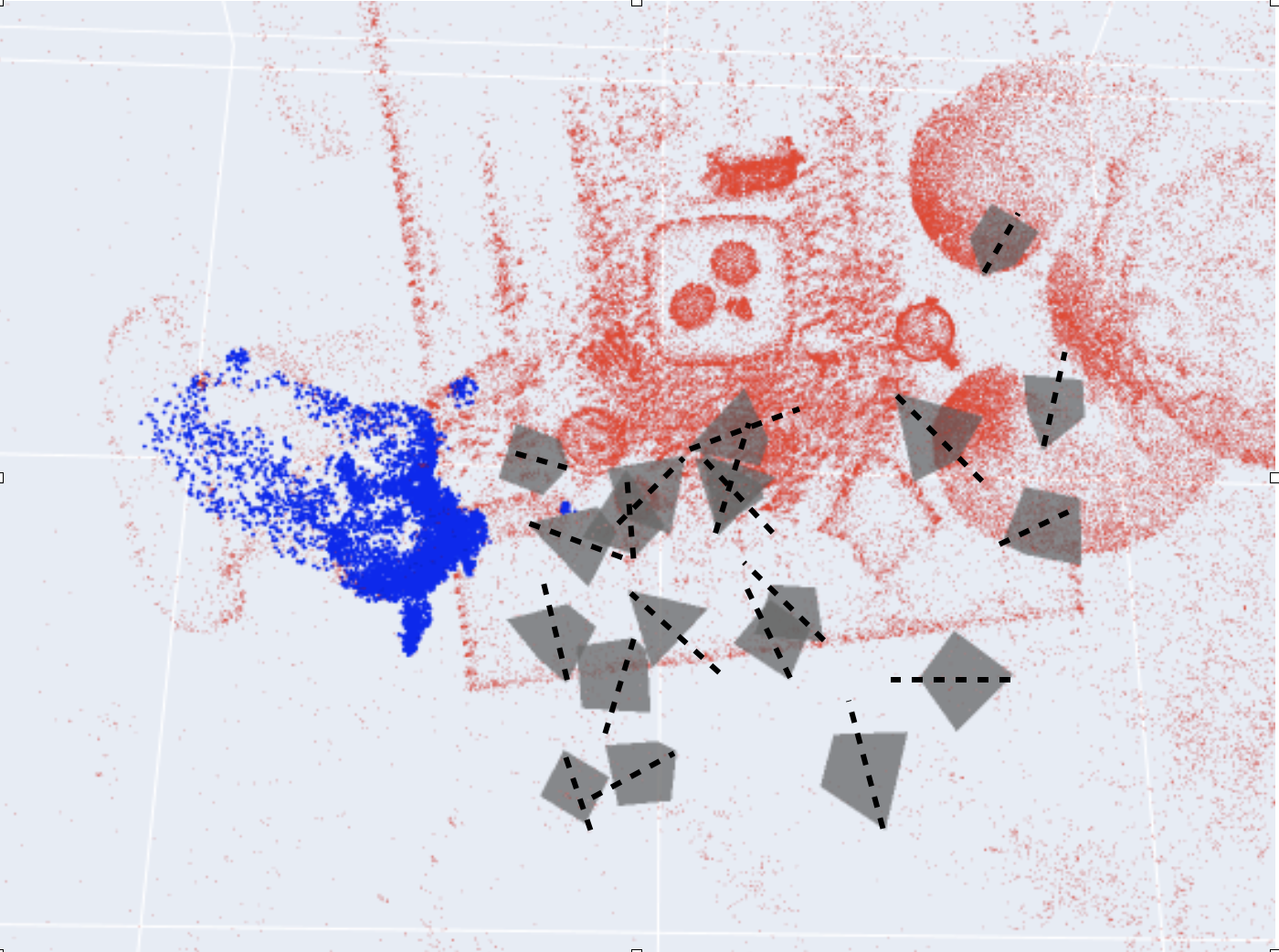}
    }%
    %\hfill
    \subfigure[Pose after SGLD]{%
        \includegraphics[width=0.48\linewidth]{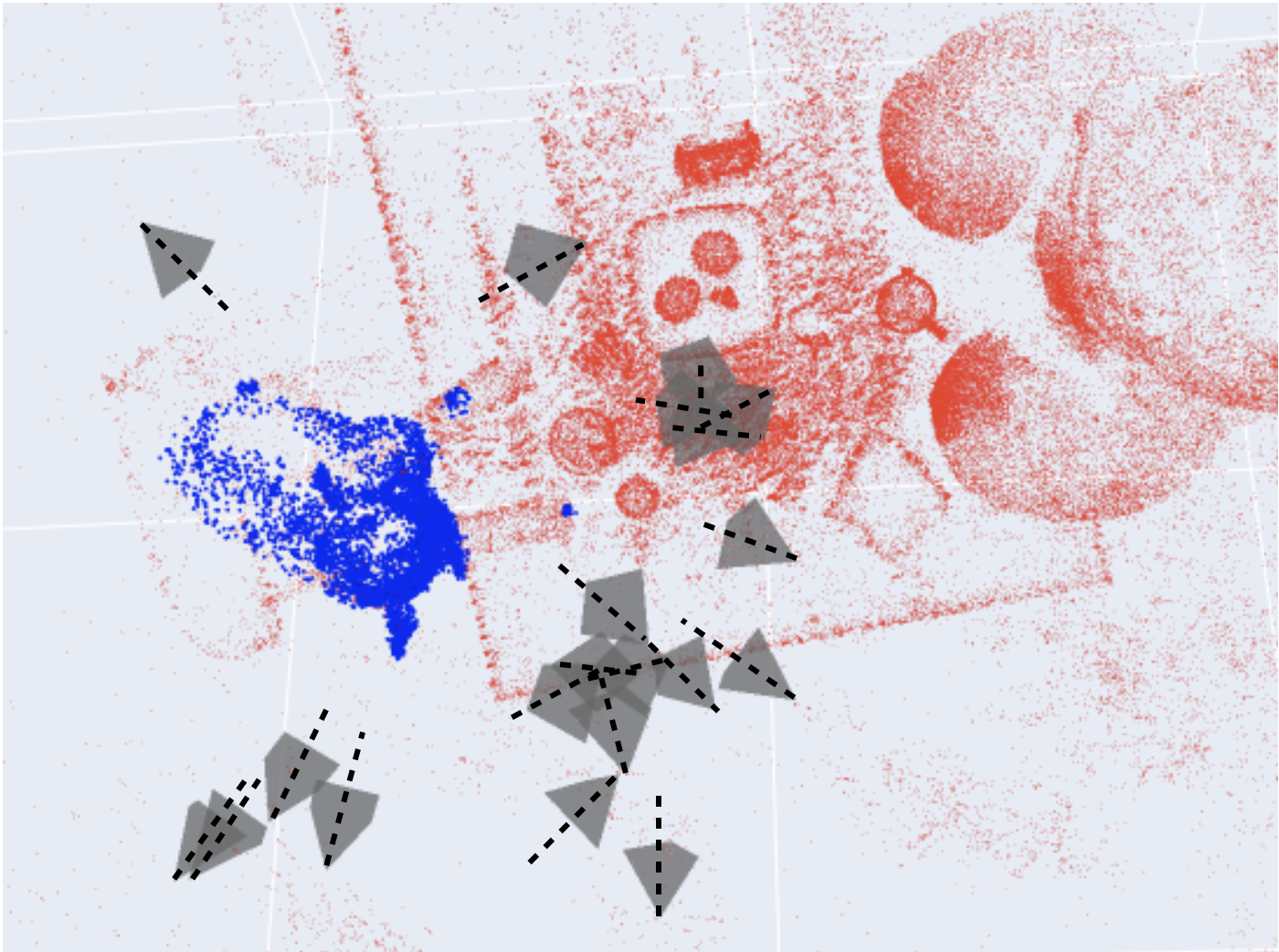}%
    }%

    \caption{A batch of camera poses before and after SGLD process. In (a), the initialized camera poses at iteration 0. In (b), optimized poses after 200 iterations of SGLD, where the cameras are pointing towards the target object (in blue).}
    \label{fig:SGLD_single_pose} 
\end{figure}

We generated 20 random poses and run SLGD to optimize the their poses. As shown in Fig. \ref{fig:SGLD_single_pose}. Rather than converging to a single deterministic solution, SGLD enables a principled approximation of the posterior distribution of the optimized camera pose. This forms a ring-shaped array pointing to the object. The wide spread in camera posterior distribution reflects that the ring-shaped low-cost region has multiple local minima, leading to different optimization results using randomized initial conditions. 

\subsection{Trajectory Optimization}
\label{subsec:traj_opt}
\begin{figure}
    \centering
    \includegraphics[width=0.7\linewidth]{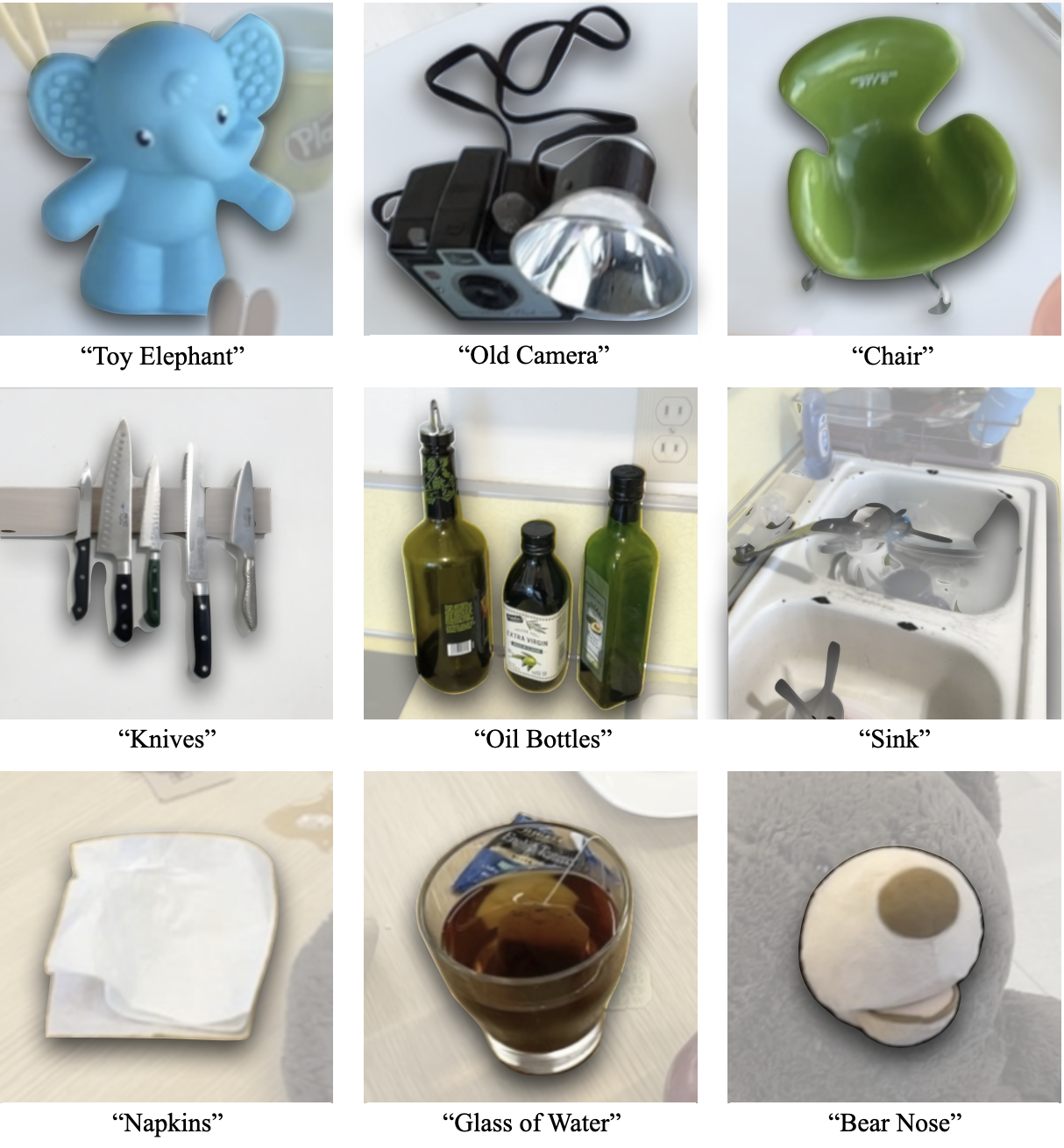}
    \caption{Gaussian Splatting rendering of queried objects}
    \label{fig:obj} 
\end{figure}
We evaluate the qualities of trajectories obtained via SplaTraj. We compare our trajectory formulation against two other baseline trajectory representations.
The experiment follows a similar procedure as the single pose experiment in \ref{subsec:single_pose}. In each experiment, we provide text prompt instruction queries to visit 3 objects within each scene. Renderings of these objects are illustrated in \cref{fig:obj}.  During each optimization, the ADAM \cite{Kingma2015AdamAM} optimizer is run to minimize the cost manifold $L^{q_i}$ for 200 iterations,  the TCE, TRE, and IoU scores are then collected from the optimized trajectories using different trajectory representations. 

\textbf{Baselines:}
 We evaluate the performance of SplaTraj against the other two commonly used representations: waypoint basis and polynomial basis.

\subsubsection{Waypoint Representation}
The following expression defines the waypoint basis functions:
\begin{equation}
\psi_i\left( t \right) = \mathbb{I}_{\{t \in T^w_j\}}, j = 1,2, \ldots m,
\end{equation}
where $\mathbb{I}_{\{A\}}$ is the indicator of event A, indicating that, the camera pose is fixed each waypoint time interval $T^w_j = [\frac{j-1}{m}, \frac{j}{m}]$. In total, we select $m=100$ waypoints in each trajectory. Waypoint representation provides a simple and deterministic method to create a piece-wise constant representation of the camera's pose over time. 

\subsubsection{Polynomial Representation}

We define our trajectory as a third-order polynomial of time. The polynomial basis function can be defined as 

\begin{equation}
\psi(t) = [t^0,t^1,t^2,\ldots,t^{N-1}]^{\top},
\end{equation}

where \( t \) is the input variable and \( N \) is the number of the max order. In our experiment, we selected $N = 6$  to strike a balance between over-fitting and under-fitting.

\textbf{Metrics:} We evaluate the quality of the image sequence and how smoothly the motion trajectory moves through the environment reconstruction.
\textbf{Intersection of Union:} Intersection of Union (IoU) is a metric for ensuring that the camera's path is optimized to capture objects accurately and consistently. The IoU score is calculated as the ratio of the area of intersection between the predicted and ground truth bounding boxes to the area of their union. This score ranges from 0 to 1, where a score of 1 indicates a perfect overlap and a score of 0 indicates no overlap at all. \textbf{Log Dimensionless Jerk:} Log Dimensionless Jerk (LDJ) \cite{LogdimJerk} is a measure used to quantify the smoothness of a motion trajectory, by considering the third derivative of motion. We seek to ensure that the generated trajectory moves smoothly across the scene.

\begin{figure}[t]
        \subfigure[Teatime]{%
        \includegraphics[width=0.30\linewidth]{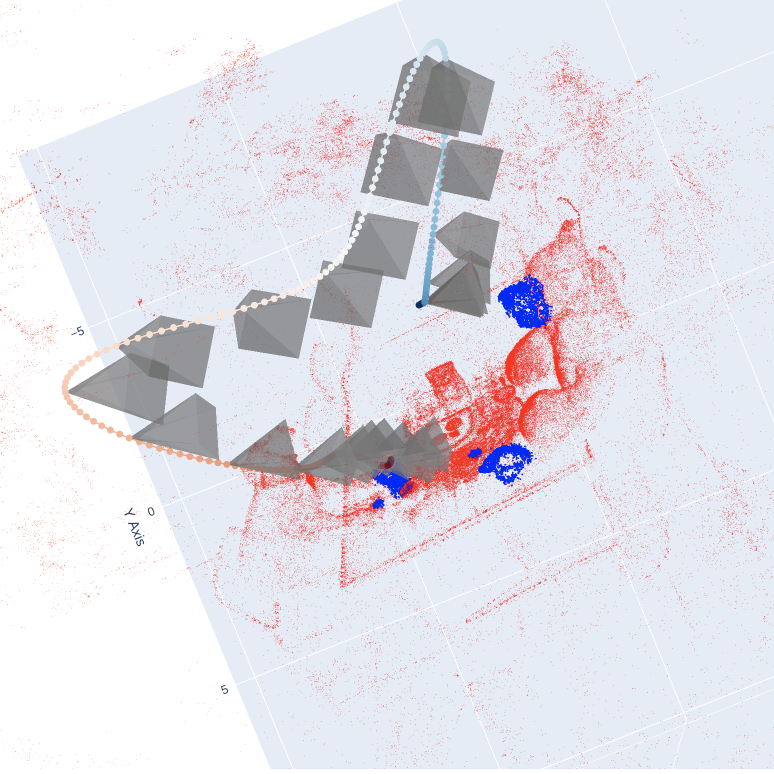}
    }%
    \hfill
    \subfigure[Kitchen]{%
        \includegraphics[width=0.30\linewidth]{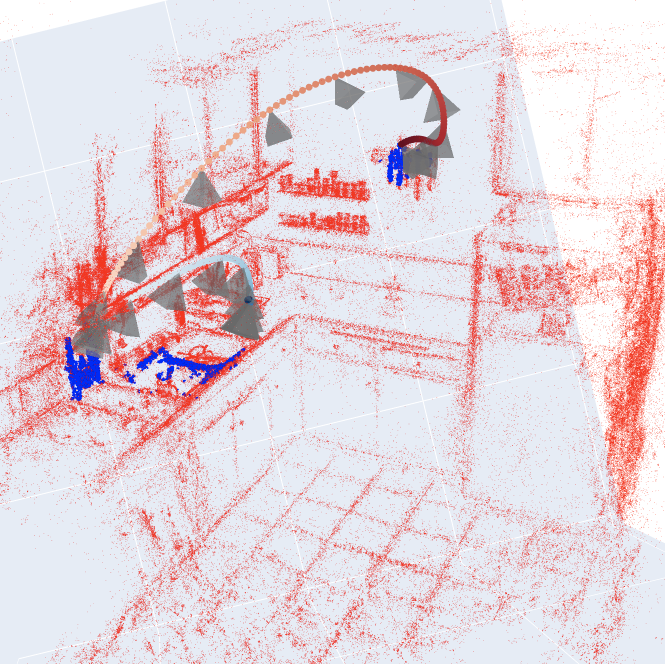}%
    }%
    \hfill
    \subfigure[Figurines]{%
        \includegraphics[width=0.30\linewidth]{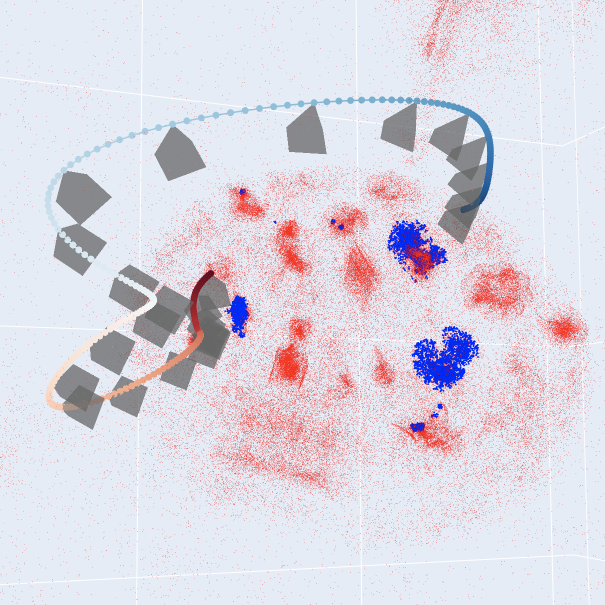}%
    }
    \hfill
    \subfigure{%
        \includegraphics[width=0.063\linewidth]{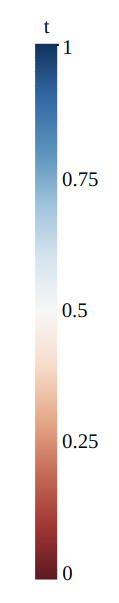}%
    }%
    \caption{We visualize the smooth trajectory the camera moves in for each of the three Gaussian Splatting models. The positions of the Gaussians in the reconstruction are illustrated in red.}
\end{figure}

\begin{table}
    \centering
    \begin{adjustbox}{width=\linewidth}
        \begin{tabular}{l|c|c|c|c}
        \toprule
        \multicolumn{2}{c|}{} & \multicolumn{3}{|c}{\textbf{Trajectory Representation}}\\
        \midrule
        \multicolumn{2}{c|}{\textbf{Metrics/Scenarios}} & \textbf{waypoints} & \textbf{polynomial} & \textbf{RBF}  \\
        \midrule 
        \midrule
        \multirow{4}{6em}{\parbox{1.3cm}{\centering \textbf{Avg. target centering error}}} & \textbf{teatime} & $0.159 \pm  0.075 $ & $\mathbf{0.143 \pm  0.120 }$ & $0.151 \pm 0.159$ \\ 
        \cmidrule{2-5}
        & \textbf{kitchen} &  $\mathbf{0.145 \pm  0.059}$ & $0.363 \pm  0.277$ & $0.228 \pm 0.198$    \\
        \cmidrule{2-5}
        & \textbf{figurines} & $ 0.169 \pm  0.072$ & $\mathbf{0.140 \pm  0.070}$ & $0.147 \pm 0.192 $       \\
        \midrule
        \midrule
        \multirow{4}{6em}{\parbox{1.3cm}{\centering \textbf{Avg. target ratio error}}} & \textbf{teatime} & $ 0.133 \pm  0.056$ &  $0.115 \pm  0.043$ & $\mathbf{0.101 \pm 0.040}$  \\ 
        \cmidrule{2-5}
        & \textbf{kitchen} & $0.125 \pm  0.044 $ & $0.158 \pm  0.042$ & $\mathbf{0.118 \pm 0.035}$    \\
        \cmidrule{2-5}
        & \textbf{figurines} &  $\mathbf{0.145 \pm  0.050}$ & $0.159 \pm  0.028$ & $0.151 \pm  0.026 $       \\
        % \cmidrule{2-5}
        
        % & \textbf{ramen} & 0 & 0 & 0      \\
        \midrule
        \midrule
        \multirow{4}{6em}{\parbox{1.3cm}{\centering \textbf{Avg. IoU }}} & \textbf{teatime} &  $0.320 \pm  0.265$ & $ 0.531 \pm  0.232$ & $\mathbf{0.626 \pm 0.178}$  \\ 
        \cmidrule{2-5}
        & \textbf{kitchen} & $0.313 \pm  0.171$ & $0.251 \pm  0.216$ & $\mathbf{0.525 \pm 0.188}$  \\
        \cmidrule{2-5}
        & \textbf{figurines} & $0.247 \pm  0.267$ & $0.325 \pm  0.223$ & $\mathbf{0.626 \pm 0.178}$       \\
        % \cmidrule{2-5}
        
        % & \textbf{ramen} & 0 & 0 & 0      \\
        \midrule
        \midrule
        \multirow{4}{6em}{\parbox{1.3cm}{\centering \textbf{Angular LDJ}}} & \textbf{Teatime} &  $15.572 \pm  0.650 $ & $7.567 \pm  0.774$ &  $ 10.229 \pm  0.599$  \\ 
        \cmidrule{2-5}
        & \textbf{kitchen} & $15.942 \pm  0.797$ & $7.909 \pm  0.965$ & $10.357 \pm  0.713$  \\
        \cmidrule{2-5}
        & \textbf{figurines} & $15.816 \pm  0.615$ & $7.513 \pm  0.832$ & $ 9.835 \pm  0.586$   \\
        % \cmidrule{2-5}
        % & \textbf{ramen} & 0 & 0 & 0      \\
        \midrule
        \midrule
        \multirow{4}{6em}{\parbox{1.3cm}{\centering \textbf{Positional LDJ}}} & \textbf{Teatime} & $15.876 \pm  0.710$ & $7.550 \pm  0.376$ &   $10.265 \pm  0.711$  \\ 
        \cmidrule{2-5}
        & \textbf{kitchen} & $16.144 \pm  0.709$ & $7.754 \pm  0.848 $ & $10.610 \pm  0.610$    \\
        \cmidrule{2-5}
        & \textbf{figurines} & $15.898 \pm  0.613$ & $7.777 \pm  0.764$ & $10.278 \pm  0.431$       \\
        % \cmidrule{2-5}
        % & \textbf{ramen} & 0 & 0 & 0      \\
        
        \bottomrule
        \end{tabular}
    \end{adjustbox}
    \vspace{0.2cm}
    \caption{Trajectory optimization results (mean $\pm$ S.D.): We compared the performance of trajectory representations (waypoints, polynomial, and RBF) across metrics and scenarios. The metrics include average target centering error, average target ratio error, average IoU, angular log dimensionless jerk (LDJ), and positional LDJ. The scenarios tested were teatime, kitchen, and figurines. The best-performing method for each metric and scenario is bold.}\label{tab:full-results}
    %\vspace{0.4cm}
    
\end{table}

\subsection{Analysis of Empirical Results} 
 % In the single pose SGDL experiment, we showed that the photogenic objective of the optimization has a ring-shaped optimal region, leading to the spread of posterior solutions, which means using different representations might lead to different optimization results. 
 We provide an analysis of our results against our baselines.

 \textbf{Waypoint:} The waypoint representation results in overfitting --- it causes discontinuities between intervals. This occurs because randomly initialized waypoints struggle to converge to a single optimal pose when multiple optima exist in the cost manifold. This can result in larger angular and positional jerks, leading to non-smooth trajectories that may cause failures in downstream robotics tasks.
    
 \textbf{Polynomial:} While polynomial representation offers smoother trajectories compared to other methods, the limited variability of the sixth-order polynomial trajectory often results in sub-optimal poses. Experiments show that the angular and positional variations in optimized polynomial trajectories are insufficient to avoid occlusions, as indicated by the low IoU score in Table \ref{tab:full-results}. Additionally, the noticeable stillness at the beginning and rapid motion near each trajectory's end, deteriorate the quality of the video.

\textbf{Radial Basis Function:} In contrast, the RBF representation produces smooth and object-oriented trajectories. The high average IoU score suggests that RBF effectively finds occlusion-avoiding poses. In contrast, the low average ratio error and centering error indicate that the trajectory is well-centered around the corresponding objects. Rendered images and 3D trajectories from each scenario, as visualized in Fig. \ref{fig:subfigures}, highlighted the trajectory generated by RBF. Compared to waypoint and polynomial representations, the RBF representation successfully provides trajectory transverse along the low-cost region defined by the optimization goal with a smooth transition between different objects.

\section{CONCLUSIONS}
\label{sec:conclusions}

In this work, we proposed SplaTraj, a framework that formalizes image sequence generation as a trajectory optimization problem. Specifically, SplaTraj enables users to specify regions and objects for the camera to visit semantically. We design costs on these 3D structures rendered to the view of a camera, which determines the desired positions within an image of the renderings. The trajectory optimizer subsequently differentiates through the rendering function and solves to obtain a continuous-time trajectory of camera poses. We empirically demonstrated that the rendering cost defined successfully leads to object-centered, properly distanced, and occlusion-avoiding rendered images, over multiple Gaussian Splatting scenes. Future avenues for research include adding further kinematics constraints into the trajectory optimization and extending SplaTraj to work in time-varying and dynamic Gaussian Splatting models.

% \renewcommand{\bibfont}{\normalfont\small}
% \printbibliography
\bibliographystyle{ieeetr}
\bibliography{reference.bib}

\begin{thebibliography}{10}

\bibitem{mildenhall2020nerf}
B.~Mildenhall, P.~P. Srinivasan, M.~Tancik, J.~T. Barron, R.~Ramamoorthi, and R.~Ng, ``Nerf: Representing scenes as neural radiance fields for view synthesis,'' in {\em ECCV}, 2020.

\bibitem{kerbl3Dgaussians}
B.~Kerbl, G.~Kopanas, T.~Leimk{\"u}hler, and G.~Drettakis, ``3d gaussian splatting for real-time radiance field rendering,'' {\em ACM Transactions on Graphics}, 2023.

\bibitem{zhang2024darkgs}
T.~Zhang, K.~Huang, W.~Zhi, and M.~Johnson-Roberson, ``Darkgs: Learning neural illumination and 3d gaussians relighting for robotic exploration in the dark,'' {\em arXiv preprint}, 2024.

\bibitem{OccupancyGridMaps}
A.~Elfes, ``Sonar-based real-world mapping and navigation,'' {\em IEEE Journal on Robotics and Automation}, 1987.

\bibitem{HM}
W.~{Zhi}, L.~{Ott}, R.~{Senanayake}, and F.~{Ramos}, ``Continuous occupancy map fusion with fast bayesian hilbert maps,'' in {\em International Conference on Robotics and Automation (ICRA)}, 2019.

\bibitem{wright2024vprism}
H.~Wright, W.~Zhi, M.~Johnson-Roberson, and T.~Hermans, ``V-prism: Probabilistic mapping of unknown tabletop scenes,'' {\em arXiv}, 2024.

\bibitem{SDF}
R.~{Malladi}, J.~A. {Sethian}, and B.~C. {Vemuri}, ``Shape modeling with front propagation: a level set approach,'' {\em IEEE Transactions on Pattern Analysis and Machine Intelligence}, 1995.

\bibitem{sptemp}
W.~{Zhi}, R.~{Senanayake}, L.~{Ott}, and F.~{Ramos}, ``Spatiotemporal learning of directional uncertainty in urban environments with kernel recurrent mixture density networks,'' {\em IEEE Robotics and Automation Letters}, 2019.

\bibitem{KTM}
W.~Zhi, L.~Ott, and F.~Ramos, ``Kernel trajectory maps for multi-modal probabilistic motion prediction,'' in {\em Conference on Robot Learning (CoRL)}, 2019.

\bibitem{mueller2022instant}
T.~M\"uller, A.~Evans, C.~Schied, and A.~Keller, ``Instant neural graphics primitives with a multiresolution hash encoding,'' {\em ACM Trans. Graph.}, 2022.

\bibitem{zhang2024recgs}
T.~Zhang, W.~Zhi, K.~Huang, J.~Mangelson, C.~Barbalata, and M.~Johnson-Roberson, ``Recgs: Removing water caustic with recurrent gaussian splatting,'' {\em arXiv preprint arXiv:2407.10318}, 2024.

\bibitem{radford2021learning}
A.~Radford, J.~W. Kim, C.~Hallacy, A.~Ramesh, G.~Goh, S.~Agarwal, G.~Sastry, A.~Askell, P.~Mishkin, J.~Clark, G.~Krueger, and I.~Sutskever, ``Learning transferable visual models from natural language supervision,'' in {\em Proceedings of the 38th International Conference on Machine Learning}, vol.~139, pp.~8748--8763, 2021.

\bibitem{qin2024langsplat3dlanguagegaussian}
M.~Qin, W.~Li, J.~Zhou, H.~Wang, and H.~Pfister, ``Langsplat: 3d language gaussian splatting,'' 2024.

\bibitem{fleishman1999automatic}
S.~Fleishman, D.~Cohen-Or, and D.~Lischinski, ``Automatic camera placement for image-based modeling,'' in {\em Proceedings of the conference on Visualization'99: celebrating ten years}, pp.~287--294, IEEE Computer Society Press, 1999.

\bibitem{albahri2017simulation}
O.~S. Albahri, ``Simulation-based optimization for camera placement in buildings,'' {\em Automation in Construction}, vol.~81, pp.~286--299, 2017.

\bibitem{bodor2005multi}
R.~Bodor, A.~Drenner, M.~Janssen, P.~Schrater, and N.~Papanikolopoulos, ``Mobile camera positioning to optimize the observability of human activity recognition tasks,'' in {\em Proceedings of the IEEE International Conference on Advanced Video and Signal Based Surveillance}, pp.~552--557, 2005.

\bibitem{Motion_planning}
S.~M. LaValle, {\em Planning Algorithms}.
\newblock USA: Cambridge University Press, 2006.

\bibitem{ompl}
I.~A. {\c{S}}ucan, M.~Moll, and L.~E. Kavraki, ``The {O}pen {M}otion {P}lanning {L}ibrary,'' {\em {IEEE} Robotics \& Automation Magazine}, pp.~72--82, December 2012.

\bibitem{pdmp}
T.~Lai, W.~Zhi, T.~Hermans, and F.~Ramos, ``Parallelised diffeomorphic sampling-based motion planning,'' in {\em Conference on Robot Learning (CoRL)}, 2021.

\bibitem{Schulman2013FindingLO}
J.~Schulman, J.~Ho, A.~X. Lee, I.~Awwal, H.~Bradlow, and P.~Abbeel, ``Finding locally optimal, collision-free trajectories with sequential convex optimization,'' in {\em Robotics: Science and Systems}, 2013.

\bibitem{CHOMP}
N.~Ratliff, M.~Zucker, J.~A. Bagnell, and S.~Srinivasa, ``Chomp: Gradient optimization techniques for efficient motion planning,'' in {\em IEEE International Conference on Robotics and Automation}, 2009.

\bibitem{Kalakrishnan2011STOMPST}
M.~Kalakrishnan, S.~Chitta, E.~Theodorou, P.~Pastor, and S.~Schaal, ``Stomp: Stochastic trajectory optimization for motion planning,'' {\em IEEE International Conference on Robotics and Automation}, 2011.

\bibitem{NMPC}
M.~Neunert, C.~de~Crousaz, F.~Furrer, M.~Kamel, F.~Farshidian, R.~Siegwart, and J.~Buchli, ``Fast nonlinear model predictive control for unified trajectory optimization and tracking,'' {\em 2016 IEEE International Conference on Robotics and Automation (ICRA)}, 2016.

\bibitem{MPPI}
G.~Williams, P.~Drews, B.~Goldfain, J.~M. Rehg, and E.~Theodorou, ``Aggressive driving with model predictive path integral control,'' in {\em IEEE International Conference on Robotics and Automation (ICRA)}, 2016.

\bibitem{RMPs}
N.~D. Ratliff, J.~Issac, D.~Kappler, S.~Birchfield, and D.~Fox, ``Riemannian motion policies,'' {\em CoRR}, 2018.

\bibitem{geoFabs}
K.~Van~Wyk, M.~Xie, A.~Li, M.~A. Rana, B.~Babich, B.~Peele, Q.~Wan, I.~Akinola, B.~Sundaralingam, D.~Fox, B.~Boots, and N.~D. Ratliff, ``Geometric fabrics: Generalizing classical mechanics to capture the physics of behavior,'' {\em IEEE Robotics and Automation Letters}, 2022.

\bibitem{GeoFab_gloabL_opt}
W.~Zhi, I.~Akinola, K.~van Wyk, N.~Ratliff, and F.~Ramos, ``Global and reactive motion generation with geometric fabric command sequences,'' in {\em IEEE International Conference on Robotics and Automation, ICRA}, 2023.

\bibitem{Fast_diff_int}
W.~Zhi, T.~Lai, L.~Ott, E.~V. Bonilla, and F.~Ramos, ``Learning efficient and robust ordinary differential equations via invertible neural networks,'' in {\em International Conference on Machine Learning, {ICML}}, 2022.

\bibitem{clip}
A.~Radford, J.~W. Kim, C.~Hallacy, A.~Ramesh, G.~Goh, S.~Agarwal, G.~Sastry, A.~Askell, P.~Mishkin, J.~Clark, G.~Krueger, and I.~Sutskever, ``Learning transferable visual models from natural language supervision,'' {\em CoRR}, 2021.

\bibitem{zheng2024gaussiangrasper3dlanguagegaussian}
Y.~Zheng, X.~Chen, Y.~Zheng, S.~Gu, R.~Yang, B.~Jin, P.~Li, C.~Zhong, Z.~Wang, L.~Liu, C.~Yang, D.~Wang, Z.~Chen, X.~Long, and M.~Wang, ``Gaussiangrasper: 3d language gaussian splatting for open-vocabulary robotic grasping,'' {\em CoRR}, 2024.

\bibitem{kerr2023lerflanguageembeddedradiance}
J.~Kerr, C.~M. Kim, K.~Goldberg, A.~Kanazawa, and M.~Tancik, ``Lerf: Language embedded radiance fields,'' 2023.

\bibitem{ester1996density}
M.~Ester, H.-P. Kriegel, J.~Sander, X.~Xu, {\em et~al.}, ``A density-based algorithm for discovering clusters in large spatial databases with noise.,'' in {\em kdd}, vol.~96, pp.~226--231, 1996.

\bibitem{kingma2017adammethodstochasticoptimization}
D.~P. Kingma and J.~Ba, ``Adam: A method for stochastic optimization,'' 2017.

\bibitem{arif2023directlyoptimizingioubounding}
M.~ul~Islam~Arif, M.~Jameel, and L.~Schmidt-Thieme, ``Directly optimizing iou for bounding box localization,'' 2023.

\bibitem{SGLD}
M.~Welling and Y.~W. Teh, ``Bayesian learning via stochastic gradient langevin dynamics,'' in {\em Proceedings of the 28th International Conference on International Conference on Machine Learning}, 2011.

\bibitem{Kingma2015AdamAM}
D.~P. Kingma and J.~Ba, ``Adam: {A} method for stochastic optimization,'' in {\em International Conference on Learning Representations}, 2015.

\bibitem{LogdimJerk}
N.~Hogan and D.~Sternad, ``Sensitivity of smoothness measures to movement duration, amplitude, and arrests,'' {\em Journal of Motor Behavior}, vol.~41, no.~6, 2009.

\end{thebibliography}
\end{document}